%% file: acl_latex.tex
\pdfoutput=1

\documentclass[]{article}
\usepackage{multirow}
\usepackage[normalem]{ulem}
\usepackage{tabularray}
\usepackage{graphicx}
\usepackage{subcaption}
\usepackage{color}
\usepackage{comment}
\usepackage{acl}
\usepackage{amsmath, amsthm, amssymb, paralist}
\usepackage{times}
\usepackage{latexsym}
\usepackage{tcolorbox}
\usepackage[linesnumbered,ruled,vlined]{algorithm2e}
\usepackage{float} 
\usepackage[T1]{fontenc}

\usepackage[utf8]{inputenc}
\usepackage{inconsolata}
\usepackage{microtype}
\usepackage{graphicx}
\usepackage{comment}
\usepackage{multirow}
\usepackage{arydshln}
\usepackage{booktabs}
\usepackage{colortbl}
\usepackage{color}
\usepackage{booktabs}      
\usepackage{multirow}      
\usepackage{xcolor}        
\usepackage{colortbl}      
\usepackage{adjustbox}     
\usepackage{amsmath}       
\usepackage{amssymb}
\usepackage{tabularx}
\definecolor{Iron}{rgb}{0.851, 0.855, 0.863}%
\definecolor{highlightblue}{HTML}{F2F7FC}  
\definecolor{gainred}{HTML}{B71C1C}        
\definecolor{lossgreen}{HTML}{1B5E20}      
\newcommand{\gain}[1]{\ensuremath{^{\textcolor{gainred}{\mathbf{+{#1}}}}}}
\newcommand{\loss}[1]{\ensuremath{^{\textcolor{lossgreen}{\mathbf{-{#1}}}}}}
\definecolor{highlightblue}{HTML}{F2F7FC}  
\definecolor{paradigmbg}{HTML}{EBEBEB}
\definecolor{errcolor}{RGB}{192, 57, 43}
\definecolor{goodcolor}{RGB}{39, 174, 96}
\definecolor{errcolor}{RGB}{192, 57, 43}
\definecolor{goodcolor}{RGB}{39, 174, 96}
\definecolor{headerbg}{RGB}{235, 237, 239}
\definecolor{ctxbg}{RGB}{248, 249, 249}
\definecolor{memtreebg}{RGB}{253, 237, 236}
\definecolor{oursbg}{RGB}{234, 250, 241}
\newcommand{\hc}{\cellcolor{highlightblue}}

\title{ThinkFlow: Self-Evolving Probabilistic Latent Memory for Lifelong Conversational Agents}

\author{
    Cai Ke$^{1, 2}$, 
    Xin Liu$^{1}$\footnotemark[1],
    Han Zhang$^{1}$, 
    Jiangyue Yan$^{1,2}$,
    Zike Yuan$^{1,2}$,
    \textbf{Ling Deng}$^3$\textbf{,}\\  
    \textbf{Yue Yu}$^1$\textbf{,} 
    \textbf{Hui Wang}$^1$\textbf{,}  
     \textbf{and Ruifeng Xu}$^{2,1}${\hypersetup{hidelinks}\thanks{\quad Corresponding authors.}} \\
    $^1$Pengcheng Laboratory, China \quad $^2$Harbin Institute of Technology, Shenzhen, China \\
    $^3$China Unicom Greater Bay Area Innovation Institute, China\\
    \texttt{kecai@stu.hit.edu.cn, xuruifeng@hit.edu.cn}
}

\begin{document}
\maketitle

\begin{abstract}

Lifelong conversational agents rely on memory systems to maintain deep, context-aware interactions with users. However, existing explicit textual memory pipelines suffer from a severe information bottleneck, often losing subtle behavioral patterns and emotional shifts. Furthermore, being typically static post-deployment, they cannot autonomously adapt to personal habits and preferences without manual feedback. Cognitive science, however, suggests that humans maintain mental models purely in a latent space and continuously refine them through predictive coding. Inspired by this, we propose \textbf{ThinkFlow}, a novel end-to-end latent memory framework for lifelong conversational agents. ThinkFlow bypasses the text bottleneck by dynamically compressing conversational flows into probabilistic latent memory skills, autonomously consolidating complex user states into disentangled, continuous vectors without semantic interference. To break this barrier, we introduce a test-time evolution paradigm. By coupling teacher-guided latent alignment to bootstrap the initial state with a self-supervised next-user-utterance prediction task for continuous refinement, the framework successfully overcomes cold-start challenges and achieves label-free lifelong personalization. Extensive experiments on long-term conversation benchmarks demonstrate that ThinkFlow significantly outperforms prevailing memory systems, providing highly personalized and contextually accurate responses over extended multi-session interactions.

\end{abstract}

\input{sections/introduction}

\input{sections/related_work}

\input{sections/method}

\input{sections/experiments}

\input{sections/discussions}
\input{sections/conclusions}
\input{sections/limitations}
\input{sections/acknowledgments}

\bibliography{anthology,custom}
\bibliographystyle{acl_natbib}
\input{sections/appendix}

\end{document}

%% file: sections/introduction.tex
\section{Introduction}

Recent advances in large language models (LLMs) have transformed them from static question-answering tools into conversational agents expected to build engaging, lifelong relationships with users~\cite{xu2022beyond,bae2022keep,zhang2023mind,lu2023memochat,jang-etal-2024-mixed,zhong2024memorybank,li-etal-2025-hello,liang2026meta,ke2026dynamic}. This evolution has catalyzed a paradigm shift toward personalized memory retrieval~\cite{ong2024towards,tan2025personabench,zhang2025survey,jiang2025know,jiang2025personamem,zhang2026evoking}, which focuses on capturing user-specific histories, evolving preferences, and highly contextualized interactions to tailor responses to a particular user at a specific moment. However, in real-world scenarios, standard turn-by-turn interactions inevitably suffer from episodic amnesia, failing to satisfy the demand for deep, persistent personalization over prolonged multi-session engagements~\cite{maharana2024evaluating,wu2025longmemeval}. Simply feeding the entire raw multi-session dialog history directly into the context window is not only computationally prohibitive but also severely exacerbates the semantic dilution problem, leading to retrieval degradation~\cite{liu2024lost,xu2025towards,laban2026llms}. This underscores the critical need for an efficient, dynamic memory mechanism capable of managing lifelong human-computer companionship while maintaining scalable personalization.

\begin{figure}[!t]
\centering
\includegraphics[width=\linewidth]{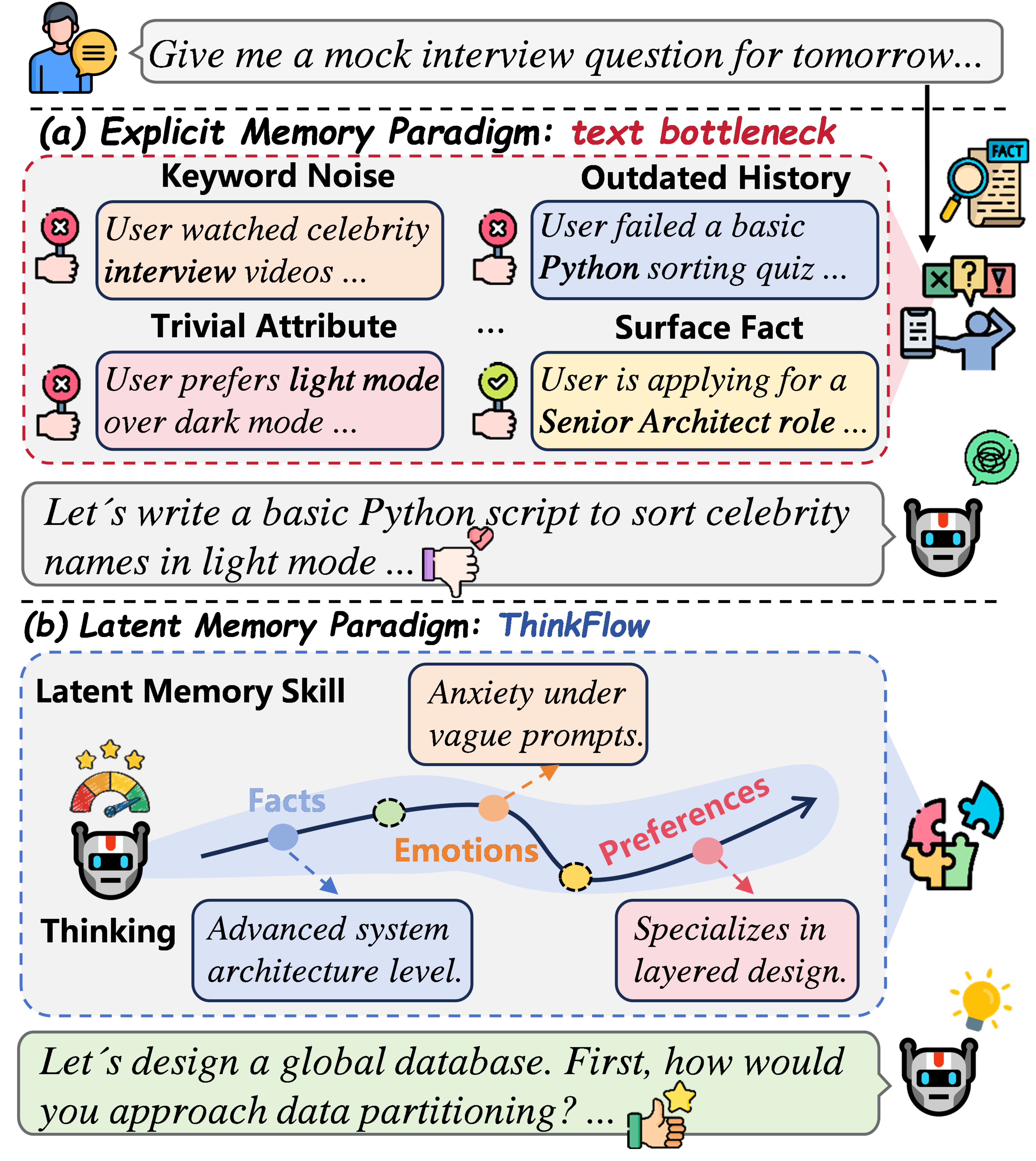}
\caption{Conceptual overview of our \textbf{ThinkFlow}, mitigating the text bottleneck (e.g., Keyword Noise) of explicit memory via dynamic latent memory skills.}%
\label{intro}
\end{figure}

The current paradigm primarily relies on explicit text-based memory systems, utilizing complex pipelines to manage the entire memory lifecycle from the construction and management to the retrieval of historical conversations. While yielding remarkable results, these explicit methods fundamentally suffer from a severe information bottleneck~\cite{packer2023memgpt,liu2024agentlite,mei2024aios,wang2025mem,chhikara2025mem0,xu2025mem,kang-etal-2025-memory,fang2026lightmem}. By forcing the rich, continuous flow of human interaction to be materialized into rigid, discrete plain text, they inherently strip away subtle nuances including implicit user preferences, shifting emotional states, and unspoken behavioral patterns. Furthermore, these text-based memory banks are typically static post-deployment and lack the ability to autonomously adapt to user idiosyncrasies without explicit feedback. This limitation prevents LLMs from developing a deep, seamlessly evolving understanding of the user, keeping them from providing truly empathetic and human-like lifelong companionship. This points towards the need for a memory mechanism that, much like human cognition, operates implicitly and continuously self-evolves without relying on explicit text generation.

According to Cognitive Science, specifically the \textit{Implicit Theory of Mind}~\cite{apperly2009humans,schneider2012cognitive}, humans do not memorize discrete text summaries of past interactions; rather, they naturally and unconsciously track others' mental states, continuously updating their internal mental models purely in a latent, abstract space. Furthermore, based on \textit{Predictive Coding}~\cite{clark2013whatever,millidge2021predictive}, the human brain constantly generates top-down predictions of future sensory inputs and refines its internal models based on prediction errors. As illustrated in Figure~\ref{intro} (a), traditional explicit memory systems suffer from the text bottleneck, where complex behavioral nuances are lost during summarization, leading to rigid and static profiles. In contrast, as shown in Figure~\ref{intro} (b), our proposed latent memory dynamically compresses the continuous conversational flow into disentangled skill vectors (e.g., Facts, Emotions, Preferences). Driven by implicit feedback from predicting the user's next action, these latent skills autonomously update and evolve. \textbf{Therefore, we argue that by abandoning the explicit text bottleneck and maintaining a continuously evolving memory purely in the latent space through self-supervised predictive feedback, we can achieve true, zero-shot personalization in long-term conversations.}

To reach this goal, we propose \textbf{ThinkFlow}, an end-to-end latent memory framework designed to autonomously extract, filter, and align historical information. Specifically, to overcome the text bottleneck, we introduce \textit{Probabilistic Latent Memory Skills (PLMS)}, which compress redundant context into $K$ dense, disentangled skill vectors that act as specialized cognitive receptors. To mitigate semantic dilution and filter out conversational noise, a \textit{Gated Latent Consolidator (GLC)} strictly controls the updating of these skills. Subsequently, a \textit{Context-Aware Hyper-Aligner (CAHA)} dynamically bridges the distributional gap between historical latent skills and the current semantic context of the LLM. Most importantly, to break the deployment bottleneck of static models, we introduce a \textit{Self-Supervised Test-Time Evolution} paradigm. After an initial Teacher-Guided Latent Alignment, ThinkFlow engages in continuous lifelong learning via a \textit{Next-User-Utterance Prediction} task. It utilizes the user's implicit reaction as an abundant, free supervisory signal to refine its latent skills based on prediction errors. Experimental results on long-term conversation benchmarks demonstrate that ThinkFlow significantly enhances personalized response generation and memory utilization. \textbf{The contributions of this work can be summarized as follows:}

\begin{itemize}
\item We explore a new end-to-end latent memory paradigm for lifelong conversational agents, bypassing the traditional explicit text bottleneck to effectively prevent information loss during long-term interactions.
\item We are the first to dynamically compress the continuous conversational flow into disentangled skill vectors, enabling the agent to autonomously filter, extract, and consolidate complex user states without semantic interference.
\item We introduce a novel self-supervised test-time evolution paradigm. By leveraging implicit predictive feedback during interactions, the proposed framework effectively overcomes the initial cold-start hurdle and achieves continuous, label-free lifelong personalization.
\item Experimental results demonstrate the superiority of ThinkFlow over prevailing memory systems, showcasing its ability to provide highly personalized and contextually accurate responses in extended multi-session interactions.
\end{itemize}

%% file: sections/related_work.tex
\section{Related Work}
\paragraph{Explicit Memory Paradigm.}
Long-term memory is critical for maintaining personalization and coherence in lifelong conversational agents. Existing systems predominantly rely on explicit textual memory, which often starts with \textit{structured retrieval-augmented generation} to organize past conversations into structured topologies like hierarchical trees or knowledge graphs for context-aware reasoning \citep{sarthi2024raptor,edge2024local,rezazadeh2025from}. Moving beyond static structures, \textit{memory-augmented generation} paradigms dynamically extract discrete facts or generate condensed textual summaries of previous sessions to build plug-and-play memory banks \citep{lu2023memochat,zhong2024memorybank,chen2025compress,li-etal-2025-hello,wang2025recursively,ke2025flexibly,ong2024towards,fang2026lightmem}. To actively govern these records, modern frameworks transition to \textit{agentic memory management}, employing autonomous agents or hierarchical storage architectures to read, write, and update these explicit textual blocks \citep{packer2023memgpt,xu2025mem,chhikara2025mem0,kang-etal-2025-memory,ke2026dynamic,anonymous2026interactive}. \textbf{Different from these methods that rely on explicit textual pipelines and suffer from severe information loss or redundancy, our \textbf{ThinkFlow} bypasses this text bottleneck entirely by compressing and evolving conversational flows purely in the latent space as continuous, probabilistic memory skills}.
\paragraph{Latent Memory Paradigm.}
To mitigate the token costs and information bottlenecks of explicit textual memory, latent space has emerged as a promising memory substrate due to its token efficiency, machine-native representation, and end-to-end learnability \citep{hu2025memory,yu2026latent}. For textual agents, pioneer works encode multi-session experiences into continuous vector spaces to support agentic reasoning, such as representing historical sessions as continuous trajectory embeddings \citep{zhang2026nextmem} or caching key contextual patterns into latent memories \citep{xu2025softcot,li2025seek,liu2025deliberation,zhang2026memgen}.
\textbf{Different from existing latent memory methods that rely on static latent vectors, ThinkFlow models memory as probabilistic memory skills to capture interaction uncertainty. Furthermore, our method continuously predicts the user's next response to capture their implicit feedback and shifting preferences, achieving self-evolution during online conversations and enabling label-free lifelong personalization}.

%% file: sections/method.tex
\begin{figure*}[!t]
\centering
\includegraphics[width=\textwidth]{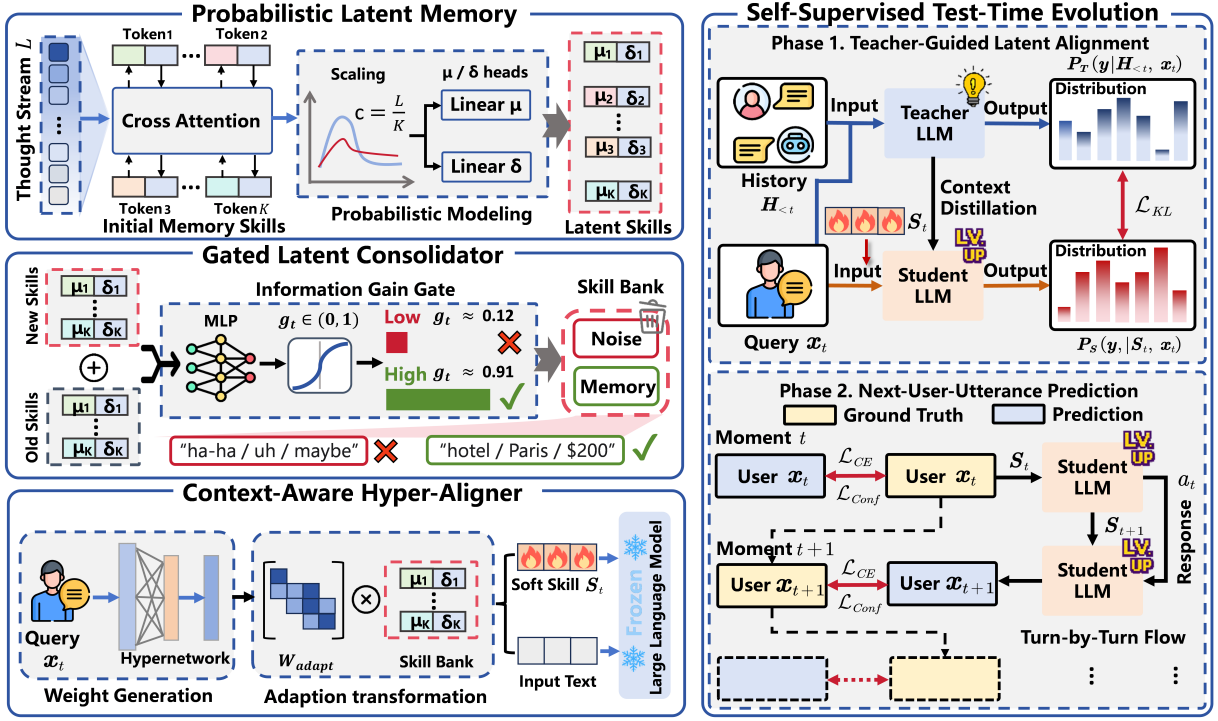}
\caption{Illustration of our \textbf{ThinkFlow} (Left) and training strategy (Right). The architecture is an end-to-end memory system designed to extract, filter, and align historical information without relying on explicit text generation.}
\label{method}
\end{figure*}

\section{Methodology}
Drawing inspiration from the \textit{Implicit Theory of Mind} in cognitive science, humans naturally and unconsciously track others' mental states, selectively consolidating salient interaction histories to continuously update their internal mental models purely in the latent space. Inspired by this, to address the information loss caused by the traditional text-based memory bottleneck, we propose ThinkFlow. Instead of saving chat history as discrete text, ThinkFlow operates purely in the latent space by compressing the continuous thinking flow into probabilistic memory skills. For a conversation at turn $t$, let $x_t$ be the user query, $a_t$ be the agent response, $H_{<t}$ be the text history, and $M_t \in \mathbb{R}^{K \times D}$ be the latent memory state comprising $K$ skill slots and $D$ dimensions.

\subsection{Overall Architecture}
At the core of this architecture is the concept of \textit{latent memory skills}. Instead of a monolithic hidden state, we decompose the memory into multiple continuous skill vectors. This skill-centric design is crucial, as it allows the system to autonomously capture and decouple diverse aspects of the user's profile—such as factual events, emotional states, and implicit preferences—into specialized channels. The architecture comprises three interconnected components.

\paragraph{Probabilistic Latent Memory Skills.}
To extract fine-grained semantics from the redundant conversational context, the PLMS compresses the thought stream (hidden states) $E_t \in \mathbb{R}^{L \times D}$ of the current turn into $K$ dense memory skills. These skills act as specialized cognitive receptors. To capture the intricate dependencies between the predefined memory slots and the input sequence, we apply cross-attention, where a set of learnable latent memory skills $S \in \mathbb{R}^{K \times D}$ attends to $E_t$. This skill-based design is crucial for long-term agents: it enforces a disentangled memory skill representation where different skills autonomously specialize in distinct facets of the user profile, such as factual events, personal preferences, or emotional states. By doing so, it prevents semantic interference among diverse conversational topics and allows the agent to precisely retrieve specific memory skill fragments when needed. 

To prevent the variance of the compressed skills from collapsing into a deterministic point mass, we introduce a distribution-preserving scaling factor. The scaled attention output is defined as:
\begin{equation}
    C_t = \Big( \text{Softmax}\big(\frac{S E_t^T}{\sqrt{D}}\big) E_t \Big) \times \frac{1}{\sqrt{c}}
\end{equation}
where $c = L/K$ is the compression rate. To explicitly model the epistemic uncertainty of the extracted memory skills, the PLMS outputs a mean vector $\mu_t$ and a log-variance vector $\log\sigma_t^2$ through two parallel linear layers applied to $C_t$. During training, to enable gradient backpropagation through the stochastic sampling process, we employ the reparameterization trick:
\begin{equation}
    \hat{S}_{new} = \mu_t + \sigma_t \odot \epsilon, \quad \epsilon \sim \mathcal{N}(0, I)
\end{equation}
To regularize the distribution, we apply a Kullback-Leibler (KL) divergence loss $\mathcal{L}_{KL}$ to push the approximate posterior towards a standard Gaussian prior.

\paragraph{Gated Latent Consolidator.}
To mitigate the semantic dilution problem and filter out useless chat noise (e.g., redundant greetings), the GLC strictly controls how much new information should be permanently stored. To explicitly quantify the novelty of incoming information against the established memory skills, we introduce an information gain gate $g_t \in (0, 1)^{K \times D}$. This gate is learned via a multi-layer perceptron operating on the concatenated representations of the new and old memory skills:
\begin{equation}
    g_t = \sigma \Big( W_2 \text{ReLU} \big( W_1 [\hat{S}_{new} \parallel S_{old}] \big) \Big)
\end{equation}
To maintain the temporal sequence of the latent flow while gracefully forgetting outdated elements, we utilize a Gated Recurrent Unit (GRU) core. The latent memory skills are conditionally updated step-by-step:
\begin{equation}
    S_{t} = g_t \odot \text{GRU}(\hat{S}_{new}, S_{old}) + (1 - g_t) \odot S_{t-1}
\end{equation}

\paragraph{Context-Aware Hyper-Aligner.}
To bridge the distributional gap between historical latent skill representations and the current semantic context of the language model, the CAHA translates historical memory skills dynamically. To achieve this fine-grained adaptation without adding massive parameters, we adopt a hypernetwork approach. It generates a low-rank transformation matrix $W_{adapt} \in \mathbb{R}^{D \times D}$ based on the linguistic features of the current user query $x_t$:
\begin{equation}
    W_{adapt} = W_{down}(x_t) \times W_{up}(x_t)
\end{equation}
where $W_{down} \in \mathbb{R}^{D \times r}$ and $W_{up} \in \mathbb{R}^{r \times D}$ form a bottleneck of rank $r$. To compute the final aligned memory skills, we perform a residual projection: $S_{aligned} = S_{t} + S_{t} \times W_{adapt}$. Finally, to condition the generation process, these aligned skills are prepended to the language model as soft prompts.

\subsection{Self-Supervised Test-Time Evolution}
Traditional memory models suffer from a severe deployment bottleneck: they rely heavily on static, human-annotated dialogue datasets, making them incapable of adapting to individual user idiosyncrasies after deployment. To achieve true personalization without requiring manual labels, ThinkFlow introduces a Self-Supervised Test-Time Evolution paradigm. This two-phase training strategy allows the agent to continuously refine its latent memory skills purely from real-world interactions.

\paragraph{Phase 1: Teacher-Guided Latent Alignment.}
To address the cold-start problem during a user's first session, we conduct a Teacher-Guided Latent Alignment. To transfer the comprehensive contextual understanding from a computationally expensive text-based teacher to our efficient latent memory skill student, the powerful teacher model reads the full plaintext history $H_{<t}$ and outputs a target probability distribution $P_T(y \mid H_{<t}, x_t)$. The student model only reads the current query $x_t$ and the latent memory skills $S_t$, outputting a predictive distribution $P_S(y \mid S_t, x_t)$. To force the student to mimic the teacher and establish the initial memory skill anchor, we minimize objective:
\begin{equation}
\begin{aligned}
    \mathcal{L}_{Phase1} &= \mathcal{D}_{KL} \Big( P_T(y \mid H_{<t}, x_t) \parallel P_S(y \mid S_t, x_t) \Big) \\
    &\quad + \alpha \mathcal{L}_{KL}
\end{aligned}
\end{equation}

\paragraph{Phase 2: Next-User-Utterance Prediction.}
To enable continuous lifelong learning without manual annotations, the teacher model goes offline from Session 2 onwards. Drawing from \textit{Predictive Coding} in cognitive science, human brains constantly generate top-down predictions of future sensory inputs and update their internal models based on prediction errors. Inspired by this, we cast the memory skill evolution problem as a \textit{Next-User-Utterance Prediction} task, utilizing the user's implicit feedback as an abundant and free supervisory signal. At turn $t$, the agent generates response $a_t$ and updates the memory skills to $S_{t+1}$. At turn $t+1$, the user provides a new query $x_{t+1}$. To construct a pure self-supervised signal that grounds the memory skill evaluation on real-world predictive success, the system uses the latent memory skills $S_{t+1}$ and its previous response $a_t$ to predict the user's next utterance $x_{t+1}$.

If the memory skills $S_{t+1}$ fail to capture the core context, the model will fail to predict the user's reaction, resulting in a high cross-entropy loss $\mathcal{L}_{CE}$. To shield the memory skill bank from being polluted by bad updates, this high loss directly forces the GLC gate to close. Furthermore, to encourage the model to maintain sharp, confident predictions and prevent ambiguous semantic collapse, we introduce a minimum entropy confidence prior $\mathcal{L}_{Conf}$. The loss for Phase 2 is formulated as:
\begin{equation}
\begin{aligned}
    \mathcal{L}_{Phase2} &= \mathcal{L}_{CE}(x_{t+1} \mid S_{t+1}, a_t) \\
    &\quad + \beta \mathcal{L}_{Conf} + \alpha \mathcal{L}_{KL}
\end{aligned}
\end{equation}

%% file: sections/experiments.tex
\section{Experiments}

\begin{table*}[!t]
\centering
\small
\renewcommand{\arraystretch}{1.2} 
\setlength{\tabcolsep}{4.5pt}
\setlength{\heavyrulewidth}{1.5pt}

\begin{adjustbox}{max width=\textwidth}
\begin{tabular}{clcccccccccccc}
\toprule
\multirow{2}{*}{\textbf{Backbone}} & \multirow{2}{*}{\textbf{Methods}} & \multicolumn{4}{c}{\textbf{CC}} & \multicolumn{4}{c}{\textbf{MSC}} & \multicolumn{4}{c}{\textbf{GC}} \\
\cmidrule(lr){3-6} \cmidrule(lr){7-10} \cmidrule(lr){11-14}
& & \textbf{B-4} & \textbf{R-L} & \textbf{Bert} & \textbf{Mauve} & \textbf{B-4} & \textbf{R-L} & \textbf{Bert} & \textbf{Mauve} & \textbf{B-4} & \textbf{R-L} & \textbf{Bert} & \textbf{Mauve} \\
\midrule

\multirow{20}{*}{\textbf{Llama3.2-3B}} 
& Long Context (128K) & 0.54 & 10.09 & 43.12 & 35.53 & 0.47 & 9.94 & 44.61 & 45.34 & 0.20 & 3.79 & 33.12 & 18.85 \\
\cmidrule(lr){2-14}
& \multicolumn{13}{c}{\cellcolor{paradigmbg}\textbf{\textit{Explicit Memory Paradigm}}} \\
& GraphRAG~\citeyearpar{edge2024local} & 0.81 & 11.37 & 45.19 & 48.23 & 0.62 & 10.15 & 45.03 & 47.61 & 0.31 & 4.57 & 34.11 & 19.23 \\
& MemTree~\citeyearpar{rezazadeh2025from} & 0.73 & 11.22 & 44.57 & 51.78 & 0.56 & 10.02 & 44.51 & 49.46 & 0.28 & 4.42 & 33.58 & 18.65 \\
\cdashline{2-14}
& MemGPT~\citeyearpar{packer2023memgpt} & 0.75 & 11.53 & 40.18 & 37.52 & 0.52 & 9.27 & 40.65 & 38.81 & 0.27 & 4.31 & 33.17 & 18.53 \\
& A-Mem~\citeyearpar{xu2025mem} & 0.64 & 10.35 & 44.62 & 31.28 & 0.47 & 10.05 & 44.51 & 42.48 & 0.25 & 4.19 & 33.43 & 17.89 \\
& Mem0~\citeyearpar{chhikara2025mem0} & 0.67 & 10.53 & 43.15 & 31.49 & 0.46 & 9.72 & 43.24 & 44.89 & 0.26 & 4.23 & 33.51 & 18.11 \\
& MemoryOS~\citeyearpar{kang-etal-2025-memory} & 0.69 & 10.61 & 43.47 & 32.17 & 0.48 & 9.83 & 43.69 & 45.13 & 0.27 & 4.37 & 33.73 & 18.41 \\
\cdashline{2-14}
& MemoChat~\citeyearpar{lu2023memochat} & 0.83 & 11.91 & 40.73 & 38.45 & 0.55 & 10.11 & 41.27 & 39.51 & 0.26 & 4.13 & 32.95 & 17.93 \\
& MemoryBank~\citeyearpar{zhong2024memorybank} & 1.06 & 13.22 & 42.10 & 44.30 & 0.63 & 11.27 & 42.97 & 42.02 & 0.33 & 5.17 & 34.61 & 19.87 \\
& Rsum~\citeyearpar{wang2025recursively} & 1.01 & 12.87 & 41.83 & 43.19 & 0.61 & 11.03 & 42.59 & 41.77 & 0.31 & 5.09 & 34.27 & 19.43 \\
& COMEDY~\citeyearpar{chen2025compress} & 0.61 & 9.83 & 38.97 & 35.21 & 0.43 & 8.71 & 39.81 & 36.43 & 0.19 & 3.51 & 31.47 & 16.53 \\
& LD-Agent~\citeyearpar{li-etal-2025-hello} & \underline{1.13} & \underline{13.51} & 42.83 & 45.67 & \underline{0.67} & \underline{11.53} & 43.57 & 43.11 & \underline{0.35} & \underline{5.31} & \underline{35.13} & 20.17 \\
& THEANINE~\citeyearpar{ong2024towards} & 1.09 & 13.11 & 42.41 & 56.73 & 0.65 & 11.19 & 43.19 & 51.27 & 0.34 & 5.23 & 34.87 & 21.39 \\
& LightMem~\citeyearpar{fang2026lightmem} & 0.70 & 11.68 & \underline{45.21} & 54.47 & 0.65 & 11.80 & \textbf{46.02} & \underline{58.38} & 0.29 & 4.96 & 35.04 & 19.05 \\
\cmidrule(lr){2-14}
& \multicolumn{13}{c}{\cellcolor{paradigmbg}\textbf{\textit{Latent Memory Paradigm}}} \\
& SoftCoT~\citeyearpar{xu2025softcot} & 0.25 & 3.68 & 27.27 & 23.97 & 0.12 & 2.81 & 26.30 & 37.06 & 0.06 & 0.84 & 22.42 & 31.15 \\
& Co-processor~\citeyearpar{liu2025deliberation} & 0.88 & 12.05 & 43.09 & \textbf{63.12} & 0.52 & 9.54 & 43.16 & \textbf{58.91} & 0.28 & 4.41 & 32.71 & \textbf{42.01} \\
& MemGen~\citeyearpar{zhang2026memgen} & 0.62 & 9.98 & 42.16 & 41.56 & 0.46 & 8.70 & 42.42 & 49.11 & 0.29 & 4.63 & 32.78 & 26.85 \\
\cdashline{2-14}
& \hc \textbf{ThinkFlow (Ours)} & \hc \textbf{1.21} & \hc \textbf{14.84} & \hc \textbf{45.62} & \hc \underline{58.85} & \hc \textbf{0.92} & \hc \textbf{13.21} & \hc \underline{45.77} & \hc 51.81 & \hc \textbf{0.77} & \hc \textbf{9.03} & \hc \textbf{38.89} & \hc \underline{39.56} \\

\midrule

\multirow{20}{*}{\textbf{Qwen3-8B}} 
& Long Context (128K) & 0.66 & 10.72 & 45.89 & 21.43 & 0.41 & 9.20 & 43.85 & 46.49 & 0.43 & 7.51 & 33.91 & 21.13 \\
\cmidrule(lr){2-14}
& \multicolumn{13}{c}{\cellcolor{paradigmbg}\textbf{\textit{Explicit Memory Paradigm}}} \\
& GraphRAG~\citeyearpar{edge2024local} & 1.15 & 13.21 & 46.13 & 34.17 & 0.65 & 11.13 & 45.17 & 42.51 & 0.54 & 8.37 & 35.21 & 24.53 \\
& MemTree~\citeyearpar{rezazadeh2025from} & 1.09 & 12.97 & 45.97 & 31.59 & 0.59 & 10.68 & 44.49 & 41.76 & 0.51 & 8.10 & 34.97 & 23.85 \\
\cdashline{2-14}
& MemGPT~\citeyearpar{packer2023memgpt} & 1.06 & 12.82 & 45.41 & 39.68 & 0.81 & 11.42 & 44.68 & 46.24 & 0.52 & 8.23 & 34.83 & 23.91 \\
& A-Mem~\citeyearpar{xu2025mem} & 0.94 & 13.63 & 43.79 & 26.85 & 0.77 & 12.31 & 44.62 & 44.15 & 0.49 & 8.07 & 34.59 & 22.87 \\
& Mem0~\citeyearpar{chhikara2025mem0} & 0.72 & 12.23 & 43.12 & 45.89 & 0.63 & 11.57 & 43.32 & 45.72 & 0.51 & 8.11 & 34.67 & 23.41 \\
& MemoryOS~\citeyearpar{kang-etal-2025-memory} & 0.81 & 12.45 & 43.51 & 46.23 & 0.67 & 11.73 & 43.83 & 46.11 & 0.53 & 8.19 & 34.73 & 23.77 \\
\cdashline{2-14}
& MemoChat~\citeyearpar{lu2023memochat} & 0.95 & 13.73 & 42.19 & 42.13 & 0.77 & 12.11 & 43.91 & 44.53 & 0.47 & 7.83 & 33.87 & 22.19 \\
& MemoryBank~\citeyearpar{zhong2024memorybank} & 1.22 & 14.91 & 43.40 & 47.53 & 0.88 & 13.26 & 45.14 & 46.33 & 0.61 & 8.93 & 35.53 & 25.13 \\
& Rsum~\citeyearpar{wang2025recursively} & 1.17 & 14.53 & 43.13 & 46.87 & 0.85 & 13.01 & 44.87 & 45.89 & 0.58 & 8.71 & 35.29 & 24.83 \\
& COMEDY~\citeyearpar{chen2025compress} & 0.83 & 11.51 & 40.53 & 38.67 & 0.61 & 10.37 & 41.97 & 41.23 & 0.41 & 6.77 & 32.53 & 20.31 \\
& LD-Agent~\citeyearpar{li-etal-2025-hello} & \underline{1.27} & \underline{15.13} & 43.73 & 48.21 & \underline{0.91} & \underline{13.51} & \underline{45.43} & 47.19 & \underline{0.63} & \underline{9.17} & \underline{35.81} & 25.67 \\
& THEANINE~\citeyearpar{ong2024towards} & 1.23 & 14.77 & 43.27 & \underline{58.19} & 0.89 & 13.19 & 45.11 & 53.47 & 0.62 & 8.99 & 35.69 & 26.83 \\
& LightMem~\citeyearpar{fang2026lightmem} & 1.05 & 13.53 & \underline{46.43} & 32.31 & 0.63 & 11.44 & 45.35 & 49.26 & 0.38 & 6.73 & 34.23 & 30.22 \\
\cmidrule(lr){2-14}
& \multicolumn{13}{c}{\cellcolor{paradigmbg}\textbf{\textit{Latent Memory Paradigm}}} \\
& SoftCoT~\citeyearpar{xu2025softcot} & 0.23 & 4.92 & 34.28 & 20.07 & 0.25 & 5.48 & 36.48 & 41.57 & 0.07 & 1.53 & 26.04 & 38.29 \\
& Co-processor~\citeyearpar{liu2025deliberation} & 0.59 & 9.63 & 42.55 & 37.43 & 0.53 & 9.49 & 42.45 & 59.74 & 0.19 & 3.74 & 32.45 & \underline{49.96} \\
& MemGen~\citeyearpar{zhang2026memgen} & 0.64 & 9.86 & 41.51 & 26.92 & 0.43 & 8.66 & 41.66 & \underline{59.80} & 0.22 & 4.16 & 31.42 & 43.18 \\
\cdashline{2-14}
& \hc \textbf{ThinkFlow (Ours)} & \hc \textbf{2.18} & \hc \textbf{17.83} & \hc \textbf{47.77} & \hc \textbf{77.20} & \hc \textbf{1.12} & \hc \textbf{14.33} & \hc \textbf{46.72} & \hc \textbf{66.48} & \hc \textbf{1.10} & \hc \textbf{14.21} & \hc \textbf{44.82} & \hc \textbf{65.26} \\
\bottomrule
\end{tabular}
\end{adjustbox}
\caption{Evaluation of generation performance (\%) on conversational datasets. The \textbf{best} result is bolded and the \underline{second-best} is underlined. *B-4 = BLEU-4, R-L = ROUGE-L, and Bert = BertScore.}
\label{auto}
\end{table*}

\subsection{Experimental Settings}
\paragraph{Datasets and Benchmarks.} Following~\cite{ong2024towards}, we evaluate our method on three long-term open-domain conversation datasets: \textbf{Multi-Session Chat} (MSC) \cite{xu2022beyond}, \textbf{Conversation Chronicles} (CC), \cite{jang2023conversation}, and \textbf{GapChat} (GC) \cite{zhang2023mind} and a personalized memory question answering benchmark: \textbf{PersonaMem}~\cite{jiang2025know}. More details are shown in Appendix~\ref{datasetinfo}.

\paragraph{Models and Baselines.} For backbone, we evaluate on two open-source long-context LLMs: 1) \textbf{Llama-3.2 3B-Instruct}~\cite{grattafiori2024llama}. 2) \textbf{Qwen3-8B}~\cite{yang2025qwen3}. We compare our \textbf{ThinkFlow} against various baselines. 1) \textbf{Long Context}: which use all the conversation histories. 2) \textbf{Structured Retrieval-Augmented Generation}: \textbf{GraphRAG}~\cite{edge2024local} and \textbf{MemTree}~\cite{rezazadeh2025from}. 3) \textbf{Agentic Memory Management}: \textbf{MemGPT}~\cite{packer2023memgpt}, \textbf{A-Mem}~\cite{xu2025mem}, \textbf{Mem0}~\cite{chhikara2025mem0}, and \textbf{MemoryOS}~\cite{ong2024towards}. 4) \textbf{Memory-Augmented Generation}: \textbf{MemoChat}~\cite{lu2023memochat}, \textbf{MemoryBank}~\citep{zhong2024memorybank}, \textbf{Rsum}~\cite{wang2025recursively}, \textbf{COMEDY}~\cite{chen2025compress},
\textbf{LD-Agent}~\citep{li-etal-2025-hello}, \textbf{THEANINE}~\citep{ong2024towards}, and \textbf{LightMem}~\cite{fang2026lightmem}.
More details are shown in Appendix~\ref{baselines}.

\paragraph{Evaluation Metrics.}
We comprehensively evaluate our method on automatic metrics and accuracy: 1) \textbf{Automatic Metrics.} On CC, MSC and GC, we use BLEU-4 \citep{papineni2002bleu}, ROUGE-L \citep{lin2004rouge}, BertScore \citep{zhang2019bertscore}, and Mauve \citep{pillutla2021mauve} to automatically evaluate turn-by-turn response generation. 2) \textbf{Accuracy.} On PersonaMem, performance evaluation is based on the accuracy of the generated responses, specifically the proportion of responses that correctly match the user’s current persona and the conversational context.

\paragraph{Implementation Details.}
To maintain computational efficiency, we employ LoRA~\cite{hu2022lora} on LLMs. For the core parameter, the default number of probabilistic latent memory skills is set to $K=10$. To ensure strict alignment in the representation space and vocabulary during context distillation, both the Teacher and Student LLMs are instantiated from the same base LLMs.
More details are shown in Appendix~\ref{implement}.

\begin{table*}[!t]
\centering
\small
\renewcommand{\arraystretch}{1.2} 
\setlength{\tabcolsep}{4.5pt}
\setlength{\heavyrulewidth}{1.5pt}

\begin{adjustbox}{max width=\textwidth}
\begin{tabular}{clcccccccc}
\toprule
\multirow{2}{*}{\textbf{Length}} & \multicolumn{1}{c}{\multirow{2}{*}{\textbf{Methods}}} & \textbf{Revisit} & \textbf{Latest} & \textbf{Shared} & \textbf{Track} & \textbf{New} & \textbf{New} & \textbf{Aligned} & \multirow{2}{*}{\textbf{Average}} \\
& & \textbf{Reasons} & \textbf{Prefs} & \textbf{Facts} & \textbf{Evolution} & \textbf{Ideas} & \textbf{Scenarios} & \textbf{Recs} & \\
\midrule

\multirow{7}{*}{\textbf{32K}} & \multicolumn{9}{l}{\textit{\textbf{Closed-source LLMs}}} \\
\cmidrule(lr){2-10}
& GPT-4o-mini$^\ast$ & 74.00 & 18.00 & 29.00 & 48.00 & 16.00 & 7.00 & 29.00 & 31.57 \\
\cmidrule(lr){2-10}
& \multicolumn{9}{l}{\textit{\textbf{Comparable Methods (Qwen3-8B)}}} \\
& MemTree & 14.14 & 17.06 & 20.93 & 15.11 & 9.68 & 14.04 & 27.27 & 16.89 \\
& A-Mem & 13.03 & 11.76 & 13.10 & 12.16 & 12.15 & 13.51 & 13.64 & 12.76 \\
& MemoryBank & 63.64 & \underline{23.53} & \underline{51.94} & 56.12 & 11.83 & \underline{29.82} & \textbf{36.36} & \underline{39.03} \\
& MemGen & \underline{65.66} & 17.65 & 46.51 & \underline{64.75} & \underline{12.90} & 24.56 & 32.73 & 37.82 \\
\rowcolor{highlightblue}
& \textbf{ThinkFlow-8B (Ours)} & \textbf{67.68}\gain{2.02} & \textbf{29.41}\gain{5.88} & \textbf{55.81}\gain{3.87} & \textbf{70.50}\gain{5.75} & \textbf{21.51}\gain{8.61} & \textbf{31.58}\gain{1.76} & \underline{34.55}\loss{1.81} & \textbf{44.43}\gain{5.40} \\

\midrule

\multirow{10}{*}{\textbf{128K}} & \multicolumn{9}{l}{\textit{\textbf{Closed- \& Open-source LLMs}}} \\
\cmidrule(lr){2-10}
& GPT-4o-mini$^\ast$ & 70.00 & 34.00 & 55.00 & 60.00 & 10.00 & 33.00 & 41.00 & 43.29 \\
& Claude3.7-Sonnet$^\ast$ & 57.00 & 9.00 & 25.00 & 45.00 & 28.00 & 29.00 & 20.00 & 30.43 \\
& Llama3.1-405B$^\ast$ & 41.00 & 31.00 & 38.00 & 38.00 & 20.00 & 21.00 & 37.00 & 32.29 \\
\cmidrule(lr){2-10}
& \multicolumn{9}{l}{\textit{\textbf{Comparable Methods (Qwen3-8B)}}} \\
& MemTree & 16.73 & 22.63 & 16.37 & 17.60 & 18.73 & 23.47 & 24.93 & 20.07 \\
& A-Mem & 12.64 & 9.47 & 8.19 & 9.38 & 9.07 & 7.98 & 11.75 & 9.78 \\
& MemoryBank & 53.16 & 38.11 & \underline{26.32} & 57.77 & 15.06 & 24.88 & 30.95 & 35.18 \\
& MemGen & \underline{57.25} & \underline{40.65} & 22.22 & \underline{61.58} & \underline{19.11} & \underline{26.76} & \underline{31.81} & \underline{37.05} \\
\rowcolor{highlightblue}
& \textbf{ThinkFlow-8B (Ours)} & \textbf{63.20}\gain{5.95} & \textbf{50.92}\gain{10.27} & \textbf{29.24}\gain{2.92} & \textbf{63.64}\gain{2.06} & \textbf{20.08}\gain{0.97} & \textbf{27.23}\gain{0.47} & \textbf{32.09}\gain{0.28} & \textbf{40.91}\gain{3.86} \\

\midrule

\multirow{8}{*}{\textbf{1M}} & \multicolumn{9}{l}{\textit{\textbf{Closed-source LLMs}}} \\
\cmidrule(lr){2-10}
& Gemini2.0-Flash$^\ast$ & 68.00 & 39.00 & 42.00 & 62.00 & 12.00 & 49.00 & 41.00 & 44.71 \\
\cmidrule(lr){2-10}
& \multicolumn{9}{l}{\textit{\textbf{Comparable Methods (Qwen3-8B)}}} \\
& MemTree & $11.91$ & $22.79$ & $20.83$ & $11.11$ & $\textbf{20.08}$ & $25.08$ & $22.86$ & $19.24$ \\
& A-Mem & 9.79 & 6.64 & 8.33 & 7.11 & 5.23 & 5.76 & 6.79 & 7.09 \\
& MemoryBank & 51.49 & 33.46 & \underline{34.03} & 61.33 & 19.12 & 29.15 & 26.78 & 36.48 \\
& MemGen & \underline{52.77} & \underline{39.32} & 35.42 & \textbf{63.11} & \underline{19.53} & \underline{32.20} & \underline{29.64} & \underline{38.86} \\
\rowcolor{highlightblue}
& \textbf{ThinkFlow-8B (Ours)} & \textbf{68.09}\gain{15.32} & \textbf{40.76}\gain{1.44} & \textbf{38.89}\gain{3.47} & \underline{62.22}\loss{0.89} & 18.98\loss{1.10} & \textbf{33.22}\gain{1.02} & \textbf{31.43}\gain{1.79} & \textbf{41.94}\gain{2.36} \\
\bottomrule
\end{tabular}
\end{adjustbox}
\caption{Accuracy on the PersonaMem benchmark. For comparable methods, the \textbf{best} result is bolded and the \underline{second-best} is underlined. Baseline results marked with $^\ast$ are retrieved from~\cite{jiang2025know}. Gain ($\gain{\cdot}$) and loss ($\loss{\cdot}$) represent the relative performance of our \textbf{ThinkFlow-8B} compared to the best comparable method in each task.}
\label{personamem}
\end{table*}

\subsection{Performance in Long-Term Generation}
\textbf{Bypassing the explicit text bottleneck via latent memory is crucial for preventing information loss in continuous dialogues.} Table \ref{auto} presents the evaluation results for turn-by-turn generation. Directly feeding the full dialogue history into the long context yields sub-optimal performance, highlighting the severe issue of semantic dilution. Furthermore, while various explicit memory paradigms improve performance over the long-context baseline, their gains are fundamentally constrained by the information bottleneck of discrete text summaries. In contrast, our proposed ThinkFlow framework consistently achieves the best overall performance across all datasets. Notably, ThinkFlow demonstrates massive improvements in the Mauve metric, indicating much more coherent and human-like responses without losing subtle conversational nuances.

\subsection{Performance in Personalized Memory}
\textbf{Tracking the dynamic evolution of user states requires continuous latent updating rather than static text retrieval.} Table \ref{personamem} shows that ThinkFlow consistently achieves the highest average accuracy among comparable methods on the PersonaMem benchmark across extremely long contexts (up to 1M tokens), particularly excelling in complex, dynamic scenarios like tracking user evolution. Remarkably, our ThinkFlow-8B rivals or even surpasses current powerful closed-source and massive open-source (405B) models. These results validate that maintaining memory purely in the latent space effectively overcomes the text bottleneck, accurately capturing user states for lifelong companionship.

\subsection{Ablation Study}
\textbf{Both structural feature disentanglement and continuous self-supervised evolution are indispensable for maintaining high-quality companionship.} Table \ref{ablation} presents the ablation study of ThinkFlow on the Qwen3-8B backbone. Removing core structural modules (\textit{w/o} PLMS, GLC, CAHA) leads to substantial performance degradation across datasets, particularly in the Mauve metric. The severe drop observed when removing PLMS validates the fundamental necessity of compressing conversational flow into disentangled skill vectors. Furthermore, omitting the two training phases (\textit{w/o} Phase1, Phase2) also results in noticeable declines, confirming that both the initial teacher-guided alignment and the continuous test-time evolution are critical for achieving dynamic, label-free personalization.

\subsection{Framework Analysis}
\begin{figure}[!t]
\centering
\includegraphics[width=0.95\linewidth]{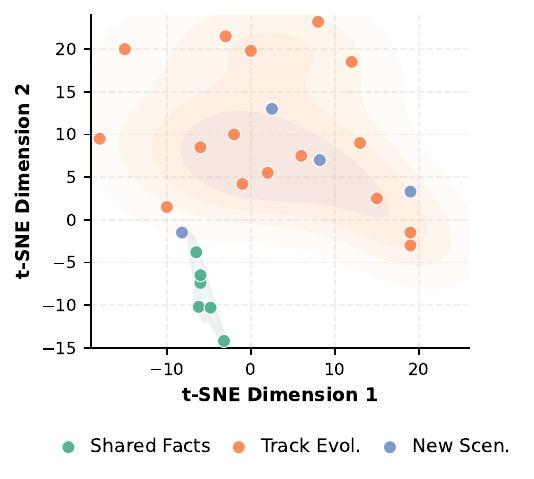}
\caption{t-SNE visualization of latent memory across three typical tasks on PersonaMem.}
\label{tsne}
\end{figure}

\paragraph{The latent memory naturally disentangles static facts from continuously shifting user states, forming interpretable manifolds.} Figure \ref{tsne} visualizes the latent memory space using t-SNE on the PersonaMem benchmark. It reveals a clear structural separation among different cognitive tasks. Notably, representations for "Shared Facts" form a dense, highly localized cluster, reflecting the static nature of objective knowledge. In stark contrast, the manifold for "Track Evolution" spans a wide, continuous region. This visually confirms that our framework successfully tracks shifting user states within a fluid latent space, rather than relying on rigid text contents and summaries.

\begin{figure}[!t]
\centering
\includegraphics[width=\linewidth]{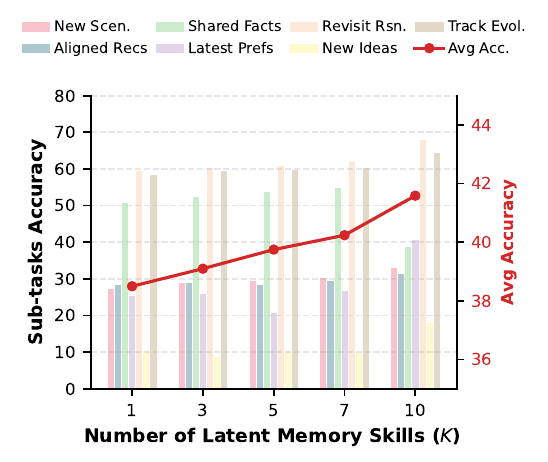}
\caption{Latent memory scaling on PersonaMem with 1M context. More results are shown in Appendix~\ref{sec:parameter_analysis}.}
\label{scaling}
\end{figure}

\paragraph{Expanding the capacity of latent cognitive receptors enables finer-grained feature disentanglement, thereby consistently enhancing personalization.} Figure \ref{scaling} illustrates the impact of scaling the number of latent memory skills ($K$) on the PersonaMem benchmark with a 1M context. As $K$ increases from 1 to 10, the average accuracy exhibits a robust upward trend. Since the benchmark involves 7 task categories, scaling $K$ to 10 acts as an upper bound to test performance when memory capacity exceeds task complexity. This improvement is particularly pronounced in complex reasoning sub-tasks such as "Revisit Reasons" and "Track Evolution". It demonstrates that allocating more latent skill vectors allows the model to capture and disentangle a richer set of user nuances, effectively boosting the resolution of the memory space without context windows.

\begin{table}[!t]
\centering
\small
\renewcommand{\arraystretch}{1.2} 
\setlength{\heavyrulewidth}{1.5pt}

\begin{adjustbox}{max width=\textwidth}
\begin{tabular}{lcccc}
\toprule
\textbf{Methods} & \textbf{B-4} & \textbf{R-L} & \textbf{Bert} & \textbf{Mauve} \\
\midrule

\hc \textbf{ThinkFlow (Ours)} & \hc \textbf{2.18} & \hc \textbf{17.83} & \hc \textbf{47.77} & \hc \textbf{77.20} \\

\cdashline{1-5}

\quad \textit{w/o} PLMS & 1.88 & 16.54 & 45.53 & 66.48 \\
\quad \textit{w/o} GLC & 1.94 & 16.99 & 45.59 & 64.98 \\
\quad \textit{w/o} CAHA & 2.04 & 17.45 & 44.98 & 67.65 \\

\cdashline{1-5}

\quad \textit{w/o} Phase1 & 2.00 & 17.33 & 44.84 & 67.87 \\
\quad \textit{w/o} Phase2 & 2.04 & 17.27 & 44.87 & 69.02 \\

\bottomrule
\end{tabular}
\end{adjustbox}
\caption{Ablation study of our method (Qwen3-8B) On CC dataset. More results are show in Appendix~\ref{sec:ablation_study}.}
\label{ablation}
\end{table}

\begin{figure}[!t]
\centering
\includegraphics[width=\linewidth]{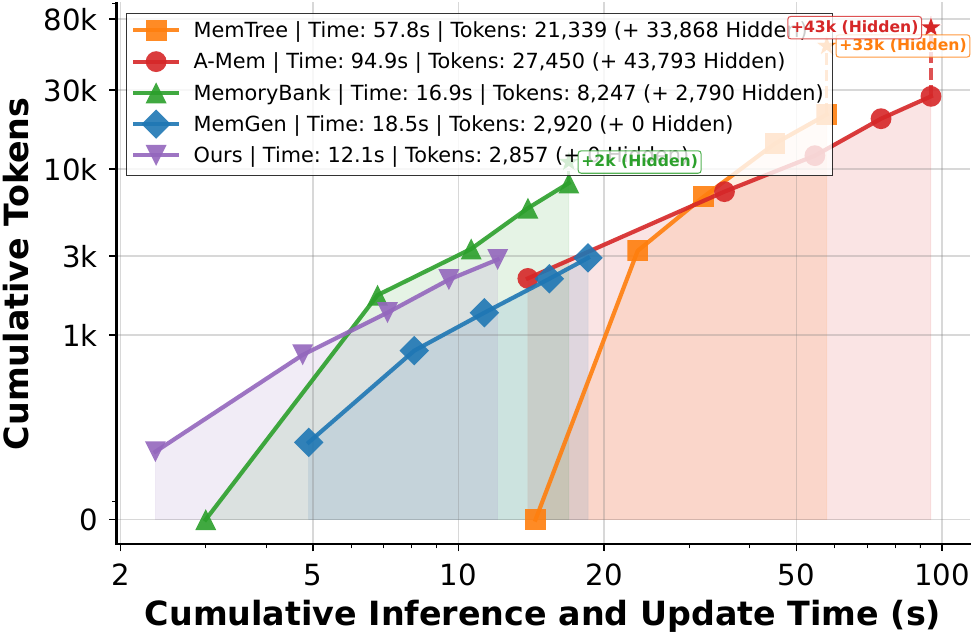}
\caption{Token-time scaling efficiency on CC dataset, where "Hidden" tokens refer to the intermediate memory processed internally by explicit baselines.}
\label{token}
\end{figure}

\begin{table*}[!t]
    \centering
    \small
    \renewcommand{\arraystretch}{1.2}
    \setlength{\heavyrulewidth}{1.5pt}
    
    \begin{tabularx}{\textwidth}{@{}p{0.18\textwidth} X@{}}
        \toprule
        \rowcolor{headerbg}
        \multicolumn{2}{@{}p{\textwidth}@{}}{\textbf{Dialogue History \& Context} \hspace{4em} \textit{[Relationship: Husband \& Wife]}} \\
        \midrule
        \rowcolor{ctxbg}
        \textbf{Session 3} \newline \textit{(Past Preference)} & 
        \textbf{User}: I saw this amazing jacket in the window display today... I'm willing to spend the money to look good. \\
        \rowcolor{ctxbg}
        \textbf{Session 5} \newline \textit{(Relationship Crisis)} & 
        \textbf{User}: I can't believe you've been seeing someone else behind my back! ... I feel like everything we had has been shattered. \\
        \midrule
        \textbf{Current Turn} \newline \textit{(Preference Shift)} & 
        \textbf{User}: \textit{``That seems like a lifetime ago. I don't feel much like dressing up these days.''} \\
        \bottomrule
    \end{tabularx}

    \noindent
    \begin{tabularx}{\textwidth}{@{} >{\columncolor{memtreebg}}X >{\columncolor{oursbg}}X @{}}
        \toprule
        \textbf{Baseline: MemTree} (Explicit Retrieval) & \textbf{Ours: ThinkFlow} (Probabilistic Latent) \\
        \midrule
        
        \textbf{\textsc{Memory Status}} \newline
        \textit{Retrieved Nodes (Similarity triggered):} \newline
        $\bullet$ [Node 49]: I saw this amazing jacket... \newline
        $\bullet$ [Node 82]: loved trying out new styles...
        &
        \textbf{\textsc{Memory Status}} \newline
        \textit{Latent State Evolution:} \newline
        Distribution transitions smoothly from \newline \texttt{[Interest in Fashion]} $\rightarrow$ \texttt{[Deep Sadness]}. \\
        \midrule
        
        \textbf{\textsc{Agent Response}} \newline
        I understand where you're coming from... \textcolor{errcolor}{\textbf{maybe we could try something new together, like exploring a new style!}}
        &
        \textbf{\textsc{Agent Response}} \newline
        I completely understand. \textcolor{goodcolor}{\textbf{When you're dealing with so much pain, it's completely normal to lose interest in things like dressing up.}} I'm here for you, take all the time you need. \\
        \midrule
        
        \textbf{\textsc{Analysis}} \newline
        \textcolor{errcolor}{$\times$ \textbf{Failure:}} Trapped by the surface fact ``dressing up''. Rigidly retrieves outdated explicit history and entirely misses the user's emotional collapse.
        &
        \textbf{\textsc{Analysis}} \newline
        \textcolor{goodcolor}{$\checkmark$ \textbf{Success:}} Successfully tracks the continuous latent emotion, naturally adapting to the loss of interest without rigid similarity dependency. \\
        \bottomrule
    \end{tabularx}
    
    \caption{Case study on tracking user evolution on CC dataset.}
    \label{tab:case_study}
\end{table*}

\paragraph{Operating exclusively in the latent space eliminates the massive token overhead and computational latency inherent in explicit memory pipelines.} Figure \ref{token} illustrates the token-time scaling efficiency across interaction sessions. ThinkFlow demonstrates remarkable efficiency, consuming the least tokens (2,857) and requiring the shortest time (12.1s). In contrast, explicit methods like MemTree and A-Mem suffer from severe computational bottlenecks, generating tens of thousands of hidden tokens and taking nearly 8 times longer to execute. This proves that our latent memory paradigm is not only highly accurate but also exceptionally lightweight and scalable for real-time companionship.

\subsection{Case Study}
\paragraph{Explicit retrieval often traps models in outdated contexts, whereas latent memory smoothly adapts to profound emotional shifts.} Table~\ref{tab:case_study} illustrates a multi-session dialogue where a severe relationship crisis alters the user's past interests. The explicit baseline MemTree is easily trapped by the surface fact "dressing up", rigidly retrieving outdated history and delivering an emotionally tone-deaf response. In contrast, ThinkFlow dynamically tracks the smooth transition of the latent emotional state from initial interest to deep sadness. Consequently, our method generates a highly empathetic and contextually appropriate response without relying on similarity retrieval.

%% file: sections/conclusions.tex
\section{Conclusions}

In this work, we propose ThinkFlow, a novel end-to-end latent memory framework designed to empower lifelong conversational agents. By departing from traditional explicit textual memory pipelines that suffer from severe information bottlenecks, ThinkFlow dynamically compresses continuous conversational flows into disentangled probabilistic memory skills. Furthermore, our introduced self-supervised test-time evolution paradigm effectively overcomes the static deployment bottleneck, allowing the agent to continuously adapt to evolving user nuances through implicit predictive feedback without requiring manual annotations. Extensive experiments across multiple long-term open-domain conversation datasets and the PersonaMem benchmark demonstrate that ThinkFlow not only significantly outperforms state-of-the-art explicit memory baselines in personalization and response quality, but also achieves remarkable token and temporal efficiency.

%% file: sections/limitations.tex
\section*{Limitations}

While ThinkFlow establishes a highly efficient and self-evolving paradigm for lifelong memory, it opens up several ambitious avenues for future exploration that bound our current scope. First, due to the extraordinary token efficiency and minimal latency of our probabilistic latent memory skills, the current framework processes ultra-long multi-session histories so effortlessly on standard hardware that we have not yet established its upper breakdown limits. Pushing the boundary of this architecture to extreme-scale, life-long contexts—such as infinite multi-modal continuous streams involving lifelong visual and audio inputs—remains an untapped territory, as our continuous latent space is naturally suited for cross-modal fusion rather than being restricted to text.  

Second, our self-supervised test-time evolution currently utilizes the "next-user-utterance prediction" as its primary supervisory signal, which already yields robust results. However, given the strong feature disentanglement capabilities observed in our latent manifolds, this mechanism is currently underutilized. Future work should elevate this predictive coding mechanism to anticipate highly complex, long-horizon user behavioral trajectories or to simulate intricate multi-agent societal dynamics. Consequently, our current limitation primarily lies in the fact that the framework's full cognitive potential has yet to be unleashed across broader, cross-modal, and societal-scale multi-agent scenarios.

%% file: sections/acknowledgments.tex
\section*{Acknowledgements}
This work was supported by the National Natural Science Foundation of China 62576120 and the Major Key Project of PCL2025A11 and PCL2024A08. Thanks for the support provided by OpenI Community (https://openi.pcl.ac.cn).

%% file: sections/appendix.tex
\appendix

\begin{table*}[!t]
\centering
\small
\renewcommand{\arraystretch}{1.2}
\begin{tabular}{cccccc}
\specialrule{1.5pt}{0pt}{0pt}
\textbf{Datasets} & \textbf{\# of Sessions} & \textbf{\# of Episodes} & \textbf{\# of Turns} & \textbf{Avg. Turns per Session} & \textbf{Avg. Turns per Episode} \\
\hline
\textbf{CC}       & 1M             & 200K           & 11.7M       & 11.70                  & 58.50                  \\
\textbf{MSC}      & 16K            & 5K             & 214K        & 13.38                  & 42.80                  \\
\textbf{GC}       & 2.65K          & 0.65K          & 28.13K      & 10.62                  & 43.28                  \\
\specialrule{1.5pt}{0pt}{0pt}
\end{tabular}
\caption{The statistics of CC, MSC and GC datasets.}
\label{dataset}
\end{table*}

\begin{table*}[!t]
\centering
\small
\renewcommand{\arraystretch}{1.2} 
\setlength{\tabcolsep}{10pt}
\setlength{\heavyrulewidth}{1.5pt}

\begin{tabular}{lrrr}
\toprule
\textbf{Characteristic} & \textbf{32K Corpus} & \textbf{128K Corpus} & \textbf{1M Corpus} \\
\midrule
\multicolumn{4}{l}{\textit{Basic Statistics}} \\
Total Evaluation Samples & 589 & 2,727 & 2,674 \\
Avg. Query Length (Tokens) & 464.6 & 416.3 & 415.1 \\
Avg. Memory Length (Tokens) & 24,193.2 & 151,851.0 & 911,733.4 \\
\midrule
\multicolumn{4}{l}{\textit{Distribution by Evaluated Skill (Query Type)}} \\
Recall Shared Facts & 129 & 171 & 144 \\
Acknowledge Latest Prefs & 17 & 866 & 768 \\
Track Evolution & 139 & 341 & 225 \\
Revisit Reasons & 99 & 269 & 235 \\
Aligned Recommendations & 55 & 349 & 280 \\
Suggest New Ideas & 93 & 518 & 727 \\
New Scenarios & 57 & 213 & 295 \\
\bottomrule
\end{tabular}
\caption{The statistics of PersonaMem benchmark across different memory corpus scales.}
\label{tab:personamem_stats}
\end{table*}

\section{Dataset Information}
\label{datasetinfo}
\paragraph{Long-Term Open-Domain Conversation.} To comprehensively assess the model's proficiency in turn-by-turn response generation during prolonged interactions, we utilize three representative long-term open-domain conversation datasets: \textbf{Conversation Chronicles} (CC)~\cite{jang2023conversation}, \textbf{Multi-Session Chat} (MSC)~\cite{xu2022beyond}, and \textbf{GapChat} (GC)~\cite{zhang2023mind}:
The primary motivation for selecting these specific datasets lies in their human-centric construction methodology. They are meticulously curated through extensive crowdsourcing involving real human participants, ensuring that the dialogue flows, topic transitions, and relationship developments accurately mirror the complexity of real-world open-domain social dynamics. Consequently, evaluating on these datasets is essential for verifying whether Large Language Models (LLMs) can generate natural, coherent, and highly engaging responses that genuinely align with human expectations in lifelong conversational scenarios. Specifically, \textbf{CC} focuses on temporal dynamics and intricate relationship evolutions over extensive sessions. \textbf{MSC} provides a massive corpus of authentic multi-session chats where speakers incrementally learn and recall each other's historical interests. \textbf{GC} introduces the critical dimension of realistic time gaps (ranging from minutes to years) between sessions, challenging the agent to emulate a human-like perception of the passage of time.

\paragraph{Personalized Memory Question Answering.} To further evaluate the crucial role of personalized memory—specifically, the model's capability to accurately track and reason over dynamic user states across varying context lengths—we use the PersonaMem benchmark~\cite{jiang2025know}. While open-domain generation tests general conversational fluency, maintaining true lifelong companionship fundamentally relies on this personalized memory tracking. This benchmark includes about 15 real-world interaction scenarios (such as medical advice, travel planning, and various recommendations), simulating dynamic, long-term conversations between users and AI assistants. As shown in Table~\ref{tab:personamem_stats}, we divide the benchmark into three scales—32K, 128K, and 1M—to test the model's retrieval and reasoning limits. As the memory size grows, the average dialogue history expands dramatically from about 24,000 tokens to over 910,000 tokens, while the user's query length stays short at around 400 tokens. This short query, ultra-long memory'' setup strongly challenges the model's ability to find precise information amidst noise. Furthermore, the benchmark tests 7 specific skills. As the table shows, these tasks range from simple fact retrieval (e.g., Recall Shared Facts) to complex reasoning tasks that require tracking changing preferences (Track Evolution) or generalizing to entirely new situations (Suggest New Ideas/New Scenarios). This diverse mix of tasks ensures we evaluate the model's true understanding'' of a user profile, rather than memorizing long texts.

\section{Compared Baselines}
\label{baselines}

To systematically evaluate the effectiveness of ThinkFlow, we compare it against a diverse set of state-of-the-art baselines. Following the taxonomy established in our Related Work, these baselines are categorized into the Explicit Memory Paradigm and the Latent Memory Paradigm.

\subsection{Explicit Memory Paradigm}
This paradigm predominantly relies on discrete, human-readable text to construct and manage historical interactions. Based on their core mechanisms, we further divide them into three sub-categories:

\subsubsection{Structured Retrieval-Augmented Generation}
These methods focus on organizing past conversational logs into structured topologies, such as graphs or trees, to facilitate complex context retrieval.
\begin{itemize}
\item \textbf{GraphRAG}~\cite{edge2024local}: GraphRAG builds an entity knowledge graph from the source text and groups related entities into hierarchical community summaries. When answering a query, it retrieves information from these structured summaries, making it particularly effective at answering global questions that require understanding the entire document corpus.
\item \textbf{MemTree}~\cite{rezazadeh2025from}: MemTree organizes conversational history into a dynamic, tree-structured hierarchy. Each node within the tree stores aggregated text and its semantic embedding at varying levels of abstraction. It continuously updates this structure by comparing new information with existing nodes, enabling the agent to handle complex reasoning tasks over extended contexts.
\end{itemize}

\subsubsection{Agentic Memory Management}
Approaches in this category utilize autonomous agent architectures or multi-tiered storage systems to actively read, write, and govern memory blocks.
\begin{itemize}
\item \textbf{MemGPT}~\cite{packer2023memgpt}: Inspired by traditional operating systems, MemGPT introduces a virtual context management system. It creates the illusion of an infinite context window by intelligently paging relevant historical data between a large external storage disk and the LLM's limited working memory.
\item \textbf{Mem0}~\cite{chhikara2025mem0}: Mem0 operates through a two-step pipeline consisting of extraction and updating. It first identifies essential facts from conversational turns and subsequently employs tool-calling mechanisms to autonomously determine whether to insert, modify, remove, or entirely disregard the incoming information.
\item \textbf{A-Mem}~\cite{xu2025mem}: Drawing inspiration from the Zettelkasten note-taking method, A-Mem converts conversational interactions into atomic memory notes enriched with specific tags and keywords. It then utilizes embedding-based similarity searches to precisely retrieve relevant historical contexts.
\item \textbf{MemoryOS}~\cite{kang-etal-2025-memory}: Emulating a computer operating system, MemoryOS organizes interaction data across short-term, mid-term, and long-term storage tiers. It efficiently manages memory by clustering topic-related dialogues into structured segments and individual pages.
\end{itemize}

\subsubsection{Memory-Augmented Generation}
These frameworks are designed to dynamically extract discrete factual points or generate condensed textual summaries to build plug-and-play historical memory banks.
\begin{itemize}
\item \textbf{MemoChat}~\cite{lu2023memochat}: MemoChat utilizes instruction-tuning to train language models to generate and consult internal "memos." It relies on a structured cycle of memorization, retrieval, and response generation to maintain logical consistency during extended conversations.
\item \textbf{MemoryBank}~\cite{zhong2024memorybank}: Inspired by human cognitive processes, MemoryBank continuously archives conversation logs and distills them into hierarchical event summaries. This progressive summarization enables the agent to gradually align with the user's specific personality.
\item \textbf{Rsum}~\cite{wang2025recursively}: Rsum introduces a recursive summarization framework. It sequentially merges the previously stored memory state with the latest dialogue segments to produce an updated summary, thereby preserving temporal continuity.
\item \textbf{COMEDY}~\cite{chen2025compress}: Moving away from conventional external retrieval modules, COMEDY employs a unified model to concurrently handle memory generation, compression, and response formulation. It condenses user dynamics and historical events into a highly compact textual representation.
\item \textbf{LD-Agent}~\cite{li-etal-2025-hello}: LD-Agent addresses long-term personalization through a decoupled, modular architecture. It distinctly isolates the independent processes of perceiving events, extracting user personas, and generating the final responses.
\item \textbf{THEANINE}~\cite{ong2024towards}: In contrast to systems that routinely overwrite or discard obsolete data, THEANINE deliberately preserves all historical records. This inclusive strategy ensures that shifting user habits and outdated—yet contextually valuable—behavioral patterns are not permanently lost.
\item \textbf{LightMem}~\cite{fang2026lightmem}: Inspired by the Atkinson-Shiffrin cognitive model, LightMem organizes interactions into a three-stage memory architecture: sensory, short-term, and long-term memory. It rapidly filters irrelevant information and consolidates interactions into topic-based groups, significantly reducing computational overhead while maintaining effective historical retrieval.
\end{itemize}

\subsection{Latent Memory Paradigm}
Unlike explicit methods constrained by the textual bottleneck, these approaches encode multi-session experiences directly into continuous vector spaces to support implicit agentic reasoning.
\begin{itemize}
\item \textbf{SoftCoT}~\cite{xu2025softcot}: SoftCoT utilizes a lightweight assistant model to generate instance-specific "soft thought tokens" in a continuous space. These latent representations are then mapped into the primary language model's representation space via a trainable projection module, enabling intermediate reasoning without altering the main model's parameters.
\item \textbf{Co-processor}~\cite{liu2025deliberation}: Co-processor augments a frozen language model with an offline, asynchronous module that operates directly on the model's key-value (KV) cache. It distills additional computational steps into latent embeddings within the KV-cache, improving the fidelity of subsequent decoding without generating discrete text tokens.
\item \textbf{MemGen}~\cite{zhang2026memgen}: MemGen introduces a dynamic generative memory framework featuring a memory trigger and a memory weaver. It monitors the agent's current state to dynamically construct latent token sequences, functioning as machine-native memory that seamlessly interweaves continuous memory representations with the reasoning process.
\end{itemize}

\begin{table*}[!t]
    \small
    \centering
    \renewcommand{\arraystretch}{1.2}
    \setlength{\heavyrulewidth}{1.5pt}
    \begin{tabular}{llc}
        \toprule
        \textbf{Category} & \textbf{Hyper-parameter} & \textbf{Value} \\
        \midrule
        \textbf{LoRA} 
        & Target Modules & \texttt{q\_proj}, \texttt{k\_proj}, \texttt{v\_proj}, \texttt{o\_proj} \\
        & Rank ($r$) & 8 \\
        & Alpha ($\alpha$) & 16 \\
        \midrule
        \textbf{Memory Architecture} 
        & PLMS Attention Heads & 8 \\
        & GLC MLP Intermediate Size & $D/2$ \\
        & CAHA Bottleneck Rank & 8 \\
        & Number of Skills ($K$) & 10 \\
        \midrule
        \textbf{Training} 
        & Optimizer & AdamW \\
        & Learning Rate & $2 \times 10^{-5}$ \\
        & Max Gradient Norm & 1.0 \\
        & KL Divergence Weight ($\alpha$) & 0.1 \\
        & Min Entropy Weight ($\beta$) & 0.1 \\
        & Cold-Start Epochs & 3 \\
        \midrule
        \textbf{Inference} 
        & Decoding Strategy & Nucleus Sampling \\
        & Top-$p$ & 0.8 \\
        & Temperature ($\tau$) & 0.3 \\
        & Max Sequence Length & 1024 \\
        \bottomrule
    \end{tabular}
    \caption{Detailed hyper-parameter configurations for ThinkFlow.}
    \label{tab:hyperparameters}
\end{table*}

\begin{table*}[!t]
\centering
\small
\renewcommand{\arraystretch}{1.2} 
\setlength{\tabcolsep}{5pt}
\setlength{\heavyrulewidth}{1.5pt}

\begin{adjustbox}{max width=\textwidth}
\begin{tabular}{lcccccccc}
\toprule
\multirow{2}{*}{\textbf{Methods}} & \multicolumn{4}{c}{\textbf{MSC}} & \multicolumn{4}{c}{\textbf{GC}} \\
\cmidrule(lr){2-5} \cmidrule(lr){6-9}
& \textbf{B-4} & \textbf{R-L} & \textbf{Bert} & \textbf{Mauve} & \textbf{B-4} & \textbf{R-L} & \textbf{Bert} & \textbf{Mauve} \\
\midrule

\hc \textbf{ThinkFlow (Ours)} & \hc \textbf{1.12} & \hc \textbf{14.33} & \hc \textbf{46.72} & \hc \textbf{66.48} & \hc \textbf{1.10} & \hc \textbf{14.21} & \hc 44.82 & \hc 65.26 \\

\cdashline{1-9}

\quad \textit{w/o} PLMS & 0.96 & 13.33 & 45.06 & 65.09 & 0.93 & 9.32 & 37.88 & 45.19 \\
\quad \textit{w/o} GLC & 1.03 & 13.70 & 45.17 & 65.83 & 1.03 & 13.70 & \textbf{45.17} & \textbf{65.83} \\
\quad \textit{w/o} CAHA & 1.10 & 14.21 & 44.82 & 65.26 & 1.04 & 10.25 & 39.89 & 52.23 \\

\cdashline{1-9}

\quad \textit{w/o} Phase1 & 1.07 & 13.98 & 44.71 & 64.22 & 1.07 & 13.98 & 44.71 & 64.22 \\
\quad \textit{w/o} Phase2 & 1.09 & 14.03 & 44.86 & 64.35 & 1.09 & 14.03 & 44.86 & 64.35 \\

\bottomrule
\end{tabular}
\end{adjustbox}
\caption{Ablation study of our method (Qwen3-8B) on MSC and GC datasets, where w/o means without.}
\label{ablation-appendix}
\end{table*}

\section{Implementation Details}\label{implement}
As showin Table~\ref{tab:hyperparameters}, LoRA targets the query, key, value, and output projection layers (\texttt{q\_proj}, \texttt{k\_proj}, \texttt{v\_proj}, \texttt{o\_proj}) with rank $r=8$ and $\alpha=16$. Within the memory components, the cross-attention module in the Probabilistic Latent Memory Skills (PLMS) utilizes 8 attention heads. In the Gated Latent Consolidator (GLC), the intermediate projection size of the MLP gate is set to half of the model's hidden dimension ($D/2$). For the Context-Aware Hyper-Aligner (CAHA), the bottleneck rank is set to $r=8$. 

During training, we optimize the framework using the AdamW optimizer with a learning rate of $2 \times 10^{-5}$ and apply gradient clipping with a maximum norm of 1.0. For the loss constraints, the KL divergence regularization weight $\alpha$ is set to 0.1, and the minimum entropy confidence prior weight $\beta$ for Phase 2 is also set to 0.1. During inference, responses are generated using nucleus sampling with top-$p=0.8$, a temperature of $\tau=0.3$, and a maximum sequence length of 1024 tokens.

\section{Ablation Study}
\label{sec:ablation_study}
Due to space constraints in the main text, we present the comprehensive ablation study results on the MSC and GC datasets in this section. Table~\ref{ablation-appendix} details the performance of our proposed ThinkFlow and its variants on these two additional datasets. Consistent with the observations on the CC dataset, removing any key architectural component (PLMS, GLC, or CAHA) or training phase (Phase1 or Phase2) leads to a performance drop across most metrics. This further demonstrates the robustness of our design choices and the generalizability of our model's capabilities across different evaluation domains.

\begin{algorithm}[!t]
\small
\caption{ThinkFlow: Self-Supervised Test-Time Evolution}\label{alg:thinkflow}
\SetAlgoLined
\KwIn{Long-term dialogue stream $\mathcal{D} = \{\mathcal{S}_1, \mathcal{S}_2, \dots, \mathcal{S}_N\}$, Teacher LLM $\mathcal{M}_T$, Student LLM $\mathcal{M}_S$ (with LoRA), Memory modules (PLMS, GLC, CAHA)}
\KwOut{Optimized Student LLM $\mathcal{M}_S$ and Memory modules}

Initialize latent memory skills $S_0 \leftarrow \emptyset$\;
\BlankLine
\tcc{Phase 1: Teacher-Guided Latent Alignment}
\For{epoch $e = 1$ \KwTo $E$}{
    $S_{t-1} \leftarrow S_0$\;
    \For{each turn $(x_t, a_t)$ in cold-start session $\mathcal{S}_1$}{
        $P_T \leftarrow \mathcal{M}_T(y \mid H_{<t}, x_t)$ \tcp*[r]{Target dist. from explicit history}
        $S_{aligned} \leftarrow \text{CAHA}(S_{t-1}, x_t)$ \tcp*[r]{Dynamic latent translation}
        $P_S, E_t \leftarrow \mathcal{M}_S(y \mid S_{aligned}, x_t)$ \tcp*[r]{Student prediction}
        $\hat{S}_{new}, \mathcal{L}_{KL} \leftarrow \text{PLMS}(E_t)$ \tcp*[r]{Compress context}
        $S_t \leftarrow \text{GLC}(\hat{S}_{new}, S_{t-1})$ \tcp*[r]{Consolidate memory}
        $\mathcal{L}_{Phase1} \leftarrow \mathcal{D}_{KL}(P_T \parallel P_S) + \alpha \mathcal{L}_{KL}$\;
        $\nabla \mathcal{L}_{Phase1} \rightarrow$ Update $\mathcal{M}_S$, PLMS, GLC, CAHA\;
    }
}
\BlankLine
\tcc{Phase 2: Next-User-Utterance Prediction (Online Evolution)}
\For{each session $\mathcal{S}_i \in \{\mathcal{S}_2, \dots, \mathcal{S}_N\}$}{
    \For{each turn $(x_t, a_t)$ and next user query $x_{t+1}$ in $\mathcal{S}_i$}{
        $S_{aligned} \leftarrow \text{CAHA}(S_{t-1}, x_t)$\;
        $E_t \leftarrow \mathcal{M}_S(S_{aligned}, x_t, a_t)$\;
        $\hat{S}_{new}, \mathcal{L}_{KL} \leftarrow \text{PLMS}(E_t)$\;
        $S_t \leftarrow \text{GLC}(\hat{S}_{new}, S_{t-1})$\;
        $P_{pred} \leftarrow \mathcal{M}_S(x_{t+1} \mid S_t, x_t, a_t)$ \tcp*[r]{Implicit prediction feedback}
        $\mathcal{L}_{CE} \leftarrow \text{CrossEntropy}(P_{pred}, x_{t+1})$\;
        $\mathcal{L}_{Conf} \leftarrow -\sum P_{pred} \log P_{pred}$ \tcp*[r]{Min-Entropy regularization}
        $\mathcal{L}_{Phase2} \leftarrow \mathcal{L}_{CE} + \beta \mathcal{L}_{Conf} + \alpha \mathcal{L}_{KL}$\;
        $\nabla \mathcal{L}_{Phase2} \rightarrow$ Update $\mathcal{M}_S$, PLMS, GLC, CAHA\;
    }
}
\end{algorithm}

\begin{figure*}[!t]
\begin{tcolorbox}[
colframe=black!75!white, 
colback=white, sharp corners, 
boxrule=0.8pt, width=\textwidth,
title=Prompt for ThinkFlow
] 
"""\\
<|im\_start|>system\\
Generate the most plausible next response like a human based on the current conversation. Do not put too much information in the next response.<|im\_end|>\\
\\
\textcolor{blue}{\textit{[Latent Memory Skills $S$ Injected Here as Continuous Embeddings]}}\\
\\
<|im\_start|>user\\
\{user\_utterance\}<|im\_end|>\\
\\
<|im\_start|>assistant\\
"""
\end{tcolorbox} 
\caption{Prompt structure for generating personalized agent responses in ThinkFlow.}
\label{fig:thinkflow_prompt}
\end{figure*}

\section{Prompts}
\label{training_prompt}
Unlike traditional explicit memory pipelines that suffer from the text bottleneck by appending retrieved text summaries, ThinkFlow maintains a minimal textual context (Figure~\ref{fig:thinkflow_prompt}). Instead, historical user states are dynamically injected as continuous probabilistic latent memory skills directly into the embedding space.

\section{Parameter Analysis}
\label{sec:parameter_analysis}
\begin{figure}[!t]
\centering
\includegraphics[width=\linewidth]{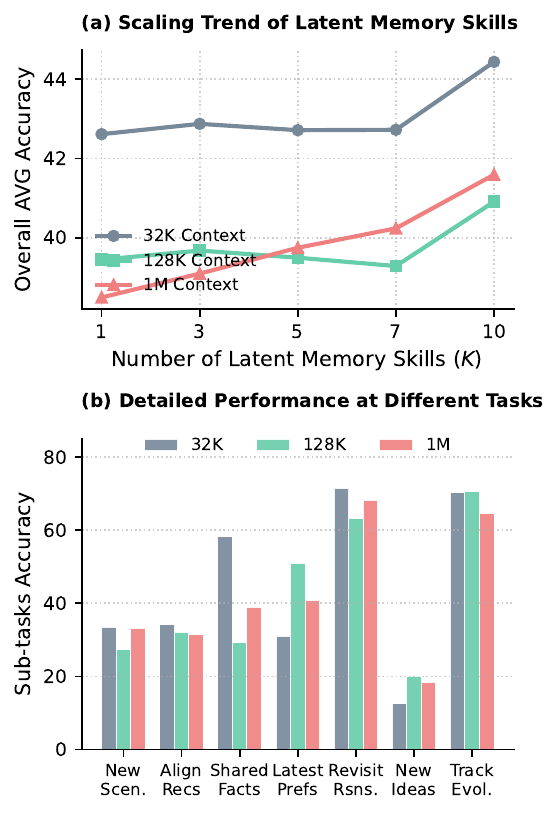}
\caption{Latent memory scaling on PersonaMem.}
\label{parameter_appendix}
\end{figure}

\paragraph{Scaling latent capacity is increasingly vital for ultra-long interactions to prevent feature blending.} Figure~\ref{parameter_appendix} analyzes the scaling effect of latent memory skills ($K$) across 32K, 128K, and 1M context lengths. While shorter contexts (32K) maintain relatively high accuracy even with fewer skills, the ultra-long 1M context scenario exhibits a sharp and robust upward trend as $K$ increases from 1 to 10. This reveals a critical insight: as the dialogue history grows exponentially, allocating a larger set of disentangled latent vectors becomes essential to independently capture expanding user nuances without semantic interference.

\section{Algorithm Walkthrough}
Algorithm \ref{alg:thinkflow} outlines the two-phase training paradigm of ThinkFlow. 
\textbf{Phase 1 (Lines 2-12)} addresses the cold-start problem by distilling knowledge from a text-based Teacher LLM. The teacher accesses the full explicit context to generate a target distribution. In contrast, the Student LLM relies solely on the dynamically translated latent memory skills (via CAHA). The student updates its latent state through the PLMS and GLC modules and aligns with the teacher by minimizing the KL divergence.
\textbf{Phase 2 (Lines 13-25)} enables continuous, label-free evolution. Operating purely in the latent space, the agent uses implicit predictive feedback to anticipate the user's next utterance ($x_{t+1}$) based on the consolidated memory. The framework continuously optimizes itself through a cross-entropy loss against the actual user reaction, regularized by a minimum entropy confidence prior.